\documentclass[letterpaper]{article} 
\usepackage[preprint]{aaai2027}  
\usepackage[hyphens]{url}  
\usepackage{graphicx} 
\usepackage{natbib}  
\usepackage{caption} 
\usepackage{algorithm}
\usepackage{algorithmic}
\usepackage[table]{xcolor}
\usepackage{amsfonts}
\usepackage{makecell}
\usepackage{siunitx}

\definecolor{datasetband}{gray}{0.93}
\definecolor{proposedrow}{gray}{0.97}

\usepackage{booktabs}
\usepackage{siunitx}

\usepackage{newfloat}
\usepackage{amsmath}
\usepackage{listings}
\DeclareCaptionStyle{ruled}{labelfont=normalfont,labelsep=colon,strut=off} 
\floatstyle{ruled}
\newfloat{listing}{tb}{lst}{}
\floatname{listing}{Listing}

\usepackage{booktabs}

\makeatletter
\renewcommand\paragraph{\@startsection{paragraph}{4}{\z@}%
  {-3pt plus -1pt minus -0.5pt}{-1em}{\normalsize\bf}}
\makeatother

\title{GaitVista: Reliability-Aware AI Measurement toward Accessible
Longitudinal Gait Assessment}

\author{
Nethmi Jayasinghe\textsuperscript{\rm 1},
Mihir Parashar\textsuperscript{\rm 2},
Amit Ranjan Trivedi\textsuperscript{\rm 1}
}
\affiliations{}

\usepackage{multirow}
\begin{document}

\maketitle

\begingroup
\renewcommand{\thefootnote}{}
\footnotetext{%
\raggedright
\footnotesize
\textsuperscript{1}Department of Electrical and Computer Engineering,
University of Illinois Chicago, Chicago, IL, USA.

\textsuperscript{2}Illinois Mathematics and Science Academy,
Aurora, IL, USA.

Correspondence: Nethmi Jayasinghe (wjayas3@uic.edu),
Amit Ranjan Trivedi (amitrt@uic.edu).
}
\endgroup

\begin{abstract}
Tracking recovery of walking function requires detecting meaningful gait
change across rehabilitation sessions, yet objective 3D measurement remains
confined to specialized motion-capture laboratories. Small camera sets and
body-worn inertial sensors broaden access, but reliability varies across joints
and time, allowing sensing failures to masquerade as patient change. We present
\textsc{GaitVista}, a reliability-aware measurement layer whose lightweight
gate assigns joint- and frame-specific visual contributions using camera
coverage, local visual quality, cross-modal disagreement, and root-motion
continuity, and exposes them for inspection. Across seven clean and degraded
sensing conditions on TotalCapture, \textsc{GaitVista} reduces average
full-body and lower-body error by \textbf{27.7\%} and \textbf{27.8\%},
attains the lowest worst-condition error among fusion methods, and reduces the
gap to a joint-frame oracle from $2.76$--$5.33$~cm for condition-blind
baselines to $1.11$~cm. On MoVi with image-derived keypoints, it is the only
deployable fusion method to improve over both unimodal streams, reducing
marker-supported error by \textbf{6.4\%} relative to the strongest learned
fusion baseline. On TotalCapture, it improves bilateral knee-flexion waveform
accuracy by \textbf{18.9\%}. Raw inertial measurements from five TotalCapture
participants show location- and time-varying magnetic disturbance, supporting
the design's reliability premise. Both benchmarks contain neurologically
healthy participants in controlled settings and retain participant-specific IMU
calibration; we therefore report progress toward accessible gait assessment,
not validated clinical deployment.
\end{abstract}

\section{Introduction}
\label{sec:introduction}

Stroke is a leading cause of long-term disability worldwide
\cite{feigin2024global}, and restoring independent ambulation is among the
outcomes patients prioritize most highly \cite{moore2022walk}. Clinicians decide
whether to continue, intensify, or change therapy by judging whether gait has
improved between sessions, and gait measures detect such change even in severely
impaired individuals \cite{henderson2022responsive}. That judgment is bounded by
the measurement available. Objective 3D gait analysis requires marker-based
capture with calibrated camera arrays, dedicated space, and trained personnel
\cite{hulleck2022present,mohan2021assessment}, so most assessment is
observational or based on one or two fixed video viewpoints. Both leave
interlimb coordination, joint excursion, balance, and left-right asymmetry only
partially observable, although asymmetry and reduced joint range are among the
most informative features of post-stroke gait
\cite{patterson2008asymmetry,xu2025asymmetry,guzik2020knee}.

A small number of consumer cameras combined with body-worn inertial measurement
units (IMUs) can be deployed in a therapy gym, community clinic, or home and
used repeatedly across sessions. Such sensing is informative, but unevenly so.
Vision recovers global body configuration and degrades under limited viewpoints
and occlusion; IMUs remain informative when body parts are hidden but are
sensitive to placement, calibration, noise, drift, and dropout
\cite{lan2022vision,garcia2023inertial}. These failures are local and
time-varying: one leg may be occluded while the torso remains visible, or an
inertial estimate may deteriorate mid-trial while the visual estimate stays
accurate.

The resulting problem is one of trust, not accuracy alone. If a system reports
that knee excursion improved by five degrees between sessions, the clinician
requires evidence that the change originates in the patient and not in the
sensing conditions. Wearable inertial systems have narrowed the accessibility
gap for several spatiotemporal measures in stroke rehabilitation
\cite{felius2022imu}, and markerless video approaches marker-based kinematics
under favorable conditions \cite{kanko2021markerless}. Repeated use outside the
laboratory nonetheless admits the conditions laboratories are built to exclude:
a therapist blocking the camera during a supported walk, a single camera in a
narrow corridor, a strap loosening over time. Under fixed or globally tuned
fusion, such events alter the measurement without indicating that sensing
quality has changed; a loosened strap resembles deterioration and an occluded
limb an apparent improvement. Longitudinal comparison is the most exposed case,
because the quantity of interest is a difference between sessions recorded under
changing conditions, a form of dataset shift identified as a barrier to reliable
clinical machine learning \cite{finlayson2021datasetshift}. A measurement system
for repeated use in uncontrolled settings must therefore determine when and
where each source of evidence is to be trusted, and record those assignments. We
treat that record as part of the reliability requirement rather than a post hoc
interpretability feature: without it, a silent sensing failure and a real change
in the patient produce similar outputs. Existing visual-inertial systems adapt
modality contributions through attention, latent state, or global confidence
\cite{trumble2017total,huang2020deepfuse,bao2022fusepose,pan2023fusing}, but
encode reliability implicitly and provide no explicit joint-time record of how a
measurement was formed \cite{kompa2021uncertainty,ghassemi2021explainability}.

\paragraph{Target setting and workflow.}
The design requirements reflect discussions with rehabilitation practitioners
who assess gait without a motion-capture laboratory, rather than a formal user
study. Three needs recur: objective measurement without dedicated facilities,
comparability across sessions, and a record of the evidence underlying each
measurement. Intended users include therapists in community rehabilitation,
outpatient follow-up, and supervised home programs. During a routine session, a
therapist records a walking trial with a small camera set and body-worn sensors;
\textsc{GaitVista} reconstructs the motion, reports assigned modality
contributions, and renders standardized virtual viewpoints independent of the
recording cameras. Holding viewpoints and derived measures fixed across visits
reduces between-session variation in downstream inspection, which motivates the
articulated Gaussian layer. The intended beneficiaries are patients for whom
repeated objective gait assessment is geographically or financially
inaccessible. The goal is to broaden access to objective movement measurement,
not to provide remote diagnosis or automated decision support.

\paragraph{Scope of the platform.} \textsc{GaitVista} comprises four components: articulated-motion reconstruction
from a small camera set and body-worn IMUs; per-joint, per-frame reliability
estimation with an inspectable record of assigned modality contributions;
standardized view generation; and rehabilitation-relevant kinematic analysis.
We evaluate the first two quantitatively, the third as a controlled downstream
case study, and the fourth through knee flexion. A lightweight visualization
prototype integrates these outputs but is not evaluated with clinicians or
patients. The objective is not pathology classification, but reliable
measurement as a prerequisite for longitudinal tracking after validation in
pathological populations.

\paragraph{Contributions.}
We develop \textsc{GaitVista}, a reliability-aware measurement layer for
accessible rehabilitation assessment:
\begin{enumerate}
    \item A formulation of accessible longitudinal measurement as a reliability
    problem rather than an accuracy problem, evaluated by regret relative to the
    stronger modality, worst-condition error, and residual gap to a per-joint,
    per-frame oracle, in addition to average error.

    \item A $1{,}794$-parameter gate producing explicit joint- and
    frame-specific modality contributions from sensing-quality, cross-modal
    disagreement, root-motion continuity, and anatomical cues, requiring neither
    condition labels nor error supervision at inference.

    \item A subject-disjoint evaluation on TotalCapture and MoVi that
    characterizes failure behavior, degradation, and oracle gap; links pose
    robustness to knee-flexion agreement and to standardized viewpoint
    generation through a fixed articulated Gaussian layer; and quantifies
    location- and time-dependent magnetic disturbance in raw inertial
    measurements from five TotalCapture participants.
\end{enumerate}

\section{Related Work}
\label{sec:related_work}

\paragraph{Accessible clinical gait assessment.}
Instrumented gait analysis provides objective measures of mobility impairment
and rehabilitation progress \cite{stebbins2023clinical,wren2020clinical}, with
marker-based capture as the reference standard, but facility, calibration, and
staffing requirements constrain its use outside laboratories
\cite{hulleck2022present}. Alternatives for post-stroke gait have been surveyed
extensively \cite{mohan2021assessment}: instrumented walkways, wearable inertial
systems, and markerless video each reduce infrastructure at some cost in
fidelity or robustness, with accuracy sensitive to camera coverage, visibility,
sensor placement, and calibration
\cite{lan2022vision,garcia2023inertial,felius2022imu,kanko2021markerless}.

\paragraph{Multimodal pose estimation and trustworthy clinical AI.}
Visual-inertial systems span dual-stream temporal fusion
\cite{trumble2017total}, wearable orientations in multi-view estimation
\cite{huang2020deepfuse}, kinematic and adaptive fusion in parametric body
models \cite{bao2022fusepose}, confidence-dependent fallback
\cite{pan2023fusing}, and diffusion, state-space, and stereo variants
\cite{pan2025diffcap,yang2026vimcan,tang2026stereo}; adaptive fusion more
broadly uses probabilistic fusion, cross-attention, uncertainty estimation, and
expert routing \cite{chen2026motion,han2024fusemoe,liu2026mixture}. We do not
claim adaptive visual-inertial fusion as new. These systems encode reliability
through latent features, temporal state, or global confidence and are judged on
average accuracy. The question here is whether explicit sensing-quality cues
support a lightweight, subject-transferable rule whose modality contributions
remain inspectable after the fact, which is the requirement imposed by
deployment shift in clinical measurement: uncommunicated uncertainty undermines
appropriate reliance \cite{kompa2021uncertainty}, post hoc explanations often
fail to support the decisions they are offered for
\cite{ghassemi2021explainability}, and distribution shift between development
and deployment is a principal failure mode \cite{finlayson2021datasetshift}.
Articulated Gaussian human representations
\cite{hu2024gauhuman,qian20243dgs,zhao2024topology,chen2024saga} assume an
accurate driving pose and are evaluated as synthesis systems. We use one for a
different purpose: as a fixed layer that regenerates identical viewpoints from
reconstructed motion, so repeated sessions are compared under matched viewing
conditions rather than the placement each session allowed.

\begin{figure*}[!t]
    \centering
    \includegraphics[width=\textwidth]{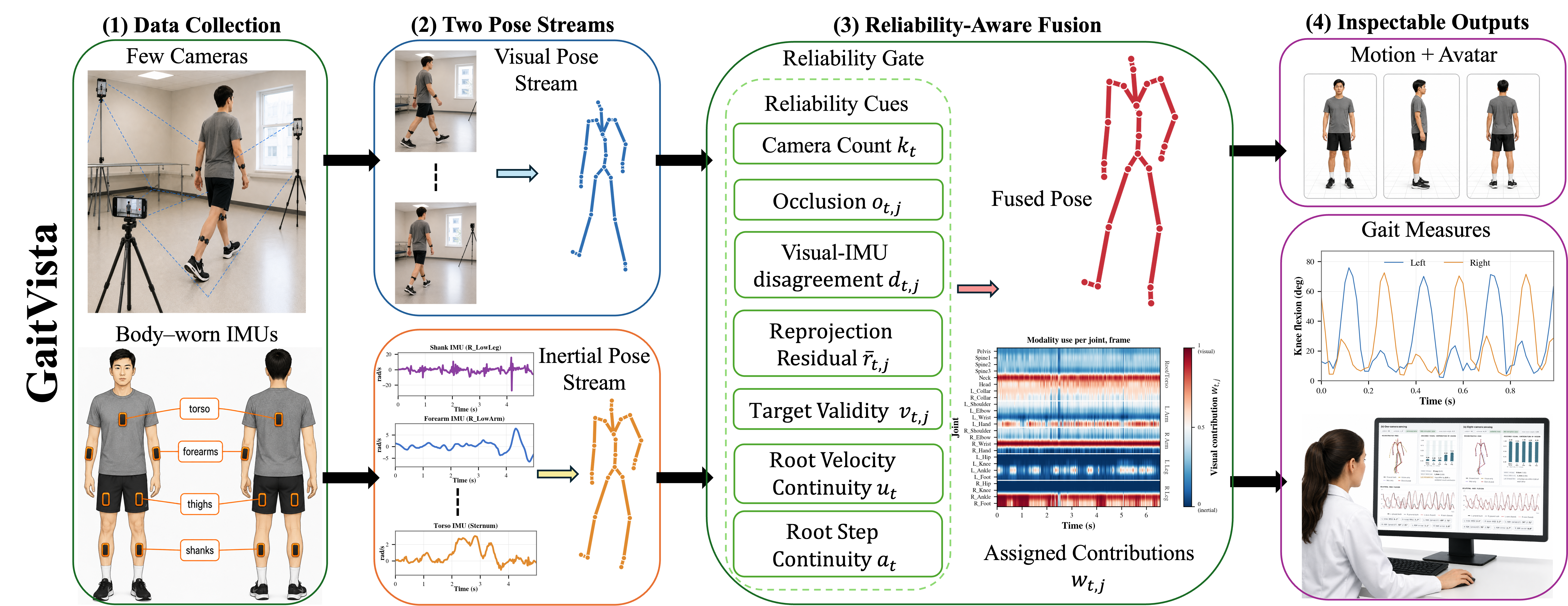}
    \vspace{-17pt}
    \caption{Overview of GaitVista. A small camera set and body-worn IMUs produce complementary visual and inertial pose estimates. Joint- and frame-specific reliability cues drive an inspectable fusion gate, yielding fused motion, assigned modality contributions, standardized avatar views, and gait measures for longitudinal review.\vspace{-15pt}}
    \label{fig:overview}
\end{figure*}

\section{Reliability-Aware Multimodal Reconstruction}
\label{sec:method}

\paragraph{Principle and Trust Criterion.}

Reliability is specific to a modality, joint, and time rather than to a
modality alone. Within one frame, vision may recover an unoccluded shoulder
accurately yet provide little evidence for a hidden leg, and an IMU may remain
reliable only until its mounting shifts, so a single global weight cannot
represent failures localized in anatomy and time.

For joint $j$ at frame $t$, an oracle selector chooses the modality closer to
ground truth. It is unavailable at inference but defines the best per-joint,
per-frame hard selection between the two streams. We therefore report, besides
average error,
\begin{equation}
G_{\mathrm{best}}
=
\min(E^{v},E^{i})-E^{f},
\label{eq:gbest}
\end{equation}
where positive values indicate improvement over both complete unimodal
estimates, together with worst-condition error and the residual gap to the
joint-frame oracle selector. These criteria test whether fusion preserves the stronger
modality, approaches locally informed selection, and stays robust as conditions
change.

The model assigns each joint and frame a visual contribution
$w_{t,j}\in[0,1]$, with values near one favoring vision and values near zero
favoring the inertial estimate. These weights form an inspectable joint-time
record of assigned modality contributions, not calibrated probabilities of
correctness.

\paragraph{Empirical Evidence for the Reliability Premise.}

\begin{figure}[!tb]
    \centering
    \includegraphics[width=0.95\columnwidth]
    {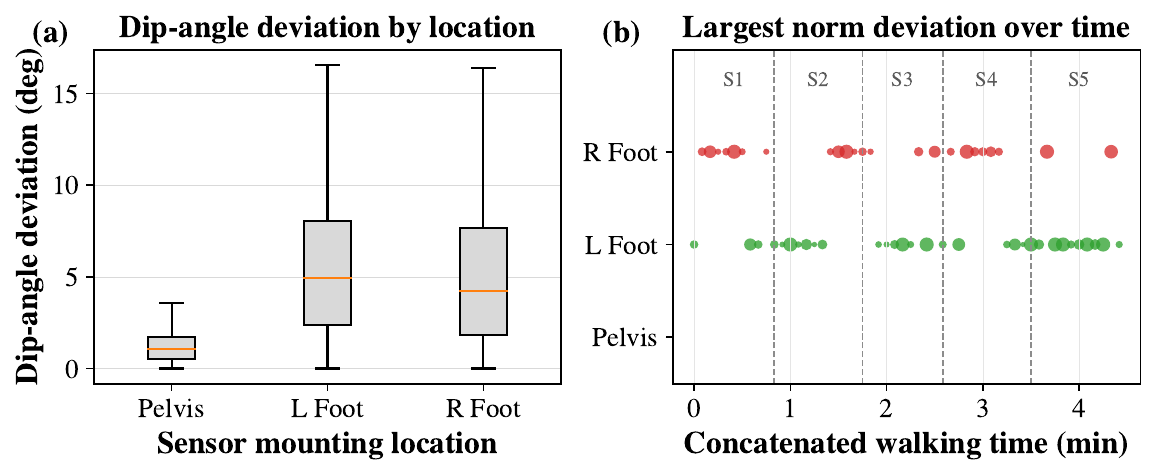}
    \vspace{-10pt}
    \caption{Location- and time-varying inertial disturbance in TotalCapture.
(a) Magnetic dip-angle deviation across sensor locations for five
\textit{walking2} participants. (b) Sensor with the largest magnetic-norm
deviation over time; dashed lines separate participants, and marker size
indicates the margin over the second-largest deviation. \vspace{-15pt} }
    \label{fig:imu_reliability}
\end{figure}

Raw inertial measurements from five TotalCapture participants show the
location- and time-dependent variation the method targets
(Fig.~\ref{fig:imu_reliability}). Across \textit{walking2} recordings,
magnetic dip-angle deviation is lower at the pelvis than at either foot-mounted
sensor, and the sensor with the largest magnetic-norm deviation changes over
time, so location-specific preferences alone cannot capture the variation.
These indicators characterize magnetic disturbance rather than pose error and
provide neither training supervision nor accuracy evidence.

\paragraph{Visual and Inertial Evidence Streams.}

At frame $t$, the visual and inertial branches produce rotations
$\mathbf{R}^{m}_{t}=\{\mathbf{R}^{m}_{t,j}\}_{j=0}^{J-1}$ for
$m\in\{v,i\}$ on a common $J=24$ joint SMPL tree. The root rotation
$\mathbf{R}^{m}_{t,0}$ is global; joints $j>0$ are parent-relative.

\noindent\textit{Visual branch.}
Given synchronized 2D joints from calibrated cameras, the visual branch
estimates the global root and local SMPL rotations by minimizing
focal-normalized multi-view reprojection error with a weak local-pose prior.
Frames are warm-started from the preceding solution, with subject shape and
camera parameters fixed. The fit also yields joint-level reprojection residuals
and target-validity indicators for reliability estimation.

\noindent\textit{Inertial branch.}
Let $\mathbf{Q}_{t,s}$ be the quaternion measured by sensor $s$ and
$\mathcal{R}(\cdot)$ its rotation-matrix conversion. After inertial-to-world
alignment $\mathbf{A}$, the world-frame measurement is
$\mathbf{M}_{t,s}=\mathbf{A}\,\mathcal{R}(\mathbf{Q}_{t,s})$. A fixed
sensor-to-joint mounting rotation is estimated by chordal averaging over a
participant-specific calibration partition:
\begin{equation}
\mathbf{C}_{s}
=
\Pi_{\mathrm{SO}(3)}
\left(
\frac{1}{|\mathcal{T}_{\mathrm{cal}}|}
\sum_{t\in\mathcal{T}_{\mathrm{cal}}}
\mathbf{M}_{t,s}^{\top}
\mathbf{R}^{\mathrm{ref}}_{t,j(s)}
\right),
\label{eq:imu_calibration}
\end{equation}
where $j(s)$ is the joint associated with sensor $s$ and
$\Pi_{\mathrm{SO}(3)}$ projects onto the rotation group. For an instrumented
joint $j$, let $s(j)$ denote its associated sensor. Its reconstructed world and
local orientations are
\begin{equation}
\widehat{\mathbf{R}}^{\,w}_{t,j}
=
\mathbf{M}_{t,s(j)}\mathbf{C}_{s(j)},
\qquad
\mathbf{R}^{i}_{t,j}
=
\left(
\widehat{\mathbf{R}}^{\,w}_{t,p(j)}
\right)^{\top}
\widehat{\mathbf{R}}^{\,w}_{t,j},
\label{eq:imu_local}
\end{equation}
for $j>0$, with
$\mathbf{R}^{i}_{t,0}=\widehat{\mathbf{R}}^{\,w}_{t,0}$.
Uninstrumented joints inherit the orientation of their associated instrumented
segment.

All pose sources share the same reference root translation, so evaluation
isolates articulated pose and root orientation rather than global trajectory
estimation. Each held-out participant retains supervised mounting calibration;
the protocol therefore tests cross-participant generalization of the fusion
model, not calibration-free deployment.

\paragraph{Joint-Wise Reliability Representation.}

The gate uses indicators of visual coverage, local measurement quality,
cross-modal disagreement, and visual temporal stability. Joint-wise
disagreement is measured after forward kinematics:
\begin{equation}
d_{t,j}
=
2\min\!\left(
\left\|
\mathbf{X}^{v}_{t,j}-\mathbf{X}^{i}_{t,j}
\right\|_{2},
0.5
\right),
\label{eq:disagreement}
\end{equation}
where positions are expressed in meters. Because both streams share the same
root translation, the cue reflects articulation and root-orientation
differences rather than global translation, and detects conflict without
ranking the streams.

Local visual quality is represented by target validity $v_{t,j}$ and the
normalized reprojection residual
$\bar r_{t,j}=\min(r_{t,j}/50,\,1)$, with missing targets assigned the
maximum. Camera count summarizes global coverage,
and a joint-level occlusion indicator marks the controlled lower-body masks
used in TotalCapture.

Reprojection error alone may not reveal temporally unstable visual poses under
sparse coverage. Let
$\Delta\mathbf{R}^{v}_{t}
=(\mathbf{R}^{v}_{t-1,0})^{\top}\mathbf{R}^{v}_{t,0}$
denote the visual root increment. We define normalized first- and second-order
continuity cues:
\begin{align}
u_t &=
\operatorname{clip}\!\left(
\frac{
d_{\mathrm{SO}(3)}
(\mathbf{I},\Delta\mathbf{R}^{v}_{t})
}{
\tau_{\mathrm{vel}}
},
0,1
\right),
\label{eq:root_velocity}\\
a_t &=
\operatorname{clip}\!\left(
\frac{
d_{\mathrm{SO}(3)}
(\Delta\mathbf{R}^{v}_{t-1},\Delta\mathbf{R}^{v}_{t})
}{
\tau_{\mathrm{acc}}
},
0,1
\right),
\label{eq:root_acceleration}
\end{align}
where $d_{\mathrm{SO}(3)}$ is the geodesic distance on
$\mathrm{SO}(3)$. These cues capture large root steps and inconsistency between
successive steps. They are broadcast across joints, with the learned joint
embedding allowing their influence to vary anatomically.

The resulting reliability vector is
\begin{equation}
\mathbf{q}_{t,j}
=
\left[
\frac{k_t}{K_{\max}},
\,o_{t,j},
\,d_{t,j},
\,\bar r_{t,j},
\,v_{t,j},
\,u_t,
\,a_t
\right]^{\top},
\label{eq:reliability_vector}
\end{equation}
where $k_t$ is the active camera count, $K_{\max}=8$, and $o_{t,j}$ is the
occlusion indicator. The cues are complementary: camera count captures global
coverage; occlusion, validity, and reprojection residual describe local visual
evidence; disagreement exposes cross-modal conflict; and root continuity
captures instability missed by frame-wise fitting. In TotalCapture,
$o_{t,j}$ is supplied by the imposed occlusion protocol and is not an observed
deployment-time signal.

\paragraph{Reliability-Adaptive Rotation Fusion.}

The reliability vector is concatenated with an eight-dimensional learned joint
embedding $\mathbf{e}_j$ and processed by a shared multilayer perceptron:
\begin{equation}
w_{t,j}
=
\sigma\!\left(
\exp(\alpha)\,
h_{\theta}\!\left(
[\mathbf{q}_{t,j};\mathbf{e}_j]
\right)
\right),
\label{eq:gate}
\end{equation}
where $w_{t,j}\in[0,1]$ is the assigned visual contribution,
$\sigma$ is the sigmoid function, and $\exp(\alpha)$ is a learned inverse
temperature. The gate has dimensions
$15\!\rightarrow\!32\!\rightarrow\!32\!\rightarrow\!1$ with ReLU
activations. Parameters are shared across joints, while $\mathbf{e}_j$ encodes
anatomical priors without requiring joint-specific gates. The model contains
$1{,}794$ trainable parameters.

Because direct matrix interpolation does not preserve $\mathrm{SO}(3)$, we
fuse rotations in the continuous six-dimensional representation
\cite{zhou2019continuity}. Let
$\phi:\mathrm{SO}(3)\rightarrow\mathbb{R}^{6}$ extract the first two columns
of a rotation matrix and
$\psi:\mathbb{R}^{6}\rightarrow\mathrm{SO}(3)$ recover a valid rotation by
Gram--Schmidt orthogonalization. The fused rotation is
\begin{equation}
\mathbf{R}^{f}_{t,j}
=
\psi\!\left(
w_{t,j}\phi(\mathbf{R}^{v}_{t,j})
+
(1-w_{t,j})\phi(\mathbf{R}^{i}_{t,j})
\right).
\label{eq:rotation_fusion}
\end{equation}
This operation is applied to the global root and all parent-relative joints.
No temporal network or post-fusion smoothing is used; temporal behavior arises
from the evidence streams and reliability cues. The weights remain available
as an inspectable joint-time record of modality contributions.

\paragraph{Condition-Balanced Learning.}

Fused rotations are mapped to 3D joints through differentiable SMPL forward
kinematics:
\begin{equation}
\mathbf{X}^{f}_{t}
=
\operatorname{FK}\!\left(
\mathbf{R}^{f}_{t};
\boldsymbol{\beta},
\mathbf{T}_{t}
\right),
\label{eq:forward_kinematics}
\end{equation}
where $\boldsymbol{\beta}$ is the fixed subject shape and $\mathbf{T}_{t}$ the
shared root translation. The gate is trained on squared position error.

Training pools sensing conditions with substantially different difficulty.
Summing raw losses would allow the most corrupted regimes to dominate
optimization and degrade typical-condition performance. We therefore normalize
each condition by the squared error of its unimodal estimate:
\begin{align}
\mathcal{L}(\theta)
&=
\sum_{c\in\mathcal{C}}
\frac{
\mathbb{E}_{(t,j)\sim c}
\left[
\left\|
\mathbf{X}^{f}_{t,j}
-
\mathbf{X}^{\mathrm{gt}}_{t,j}
\right\|_{2}^{2}
\right]
}{
b_c
},
\label{eq:balanced_loss}\\
b_c
&=
\max\!\left\{
\left[
\min\!\left(
\bar e^{v}_{c},
\bar e^{i}_{c}
\right)
\right]^2,
10^{-4}
\right\},
\label{eq:condition_normalizer}
\end{align}
where $\bar e^{v}_{c}$ and $\bar e^{i}_{c}$ are the mean visual and inertial
errors for condition $c$ on the training partition. The normalizer is computed
once per fold and held fixed. Each update draws one minibatch from every
condition, preventing severe corruptions from overwhelming cleaner regimes. Condition identity is used only to balance training and is never provided to
the gate. The model therefore learns a shared reliability rule from the
observed evidence rather than requiring the sensing regime at inference. No
auxiliary confidence supervision is used; optimization details are in the
supplement.

\paragraph{Articulated Gaussian Interpretation Layer.}

The fused motion drives a fixed articulated Gaussian avatar. Geometry,
appearance, deformation, and rendering parameters are held constant across pose
methods, so rendered differences reflect only the driving motion, and identical
viewpoints can be regenerated at every session independently of where the
cameras stood. The layer contributes no training signal and no pose refinement,
and quantitative rendering is reported only as a controlled downstream case
study.

\begin{table*}[!tb]
\centering
{
\small
\setlength{\tabcolsep}{3.8pt}
\begin{tabular}{@{}llccccc@{}}
\toprule
&
&
\multicolumn{3}{c}{MPJPE (cm) $\downarrow$}
&
$G_{\mathrm{best}}$
&
Selector gap \\
\cmidrule(lr){3-5}
Method
& Mechanism
& Avg.
& Lower
& Worst
& (cm) $\uparrow$
& (cm) $\downarrow$ \\
\midrule
Vision-only
& Unimodal
& 6.88$\pm$0.85
& 10.49$\pm$2.22
& 19.21$\pm$5.18
& --
& -- \\

IMU-only
& Unimodal
& 6.21$\pm$1.50
& 9.06$\pm$0.81
& \textbf{8.26$\pm$1.41}
& --
& -- \\

\addlinespace[2pt]
Equal-weight fusion
& Fixed blend
& 6.92$\pm$1.58
& 9.72$\pm$1.79
& 15.36$\pm$6.47
& $-2.98$
& 3.72 \\

Kalman-Condition
& Condition-tuned filter
& \underline{4.98$\pm$1.77}
& \underline{6.35$\pm$0.62}
& 11.02$\pm$5.64
& \underline{$-1.04$}
& \underline{1.78} \\

\addlinespace[2pt]
Concat+MLP
& Pose-input MLP
& 6.82$\pm$1.36
& 8.75$\pm$1.23
& 14.92$\pm$5.29
& $-2.88$
& 3.62 \\

FusePose
& Direct regression
& 6.09$\pm$1.41
& 8.58$\pm$1.00
& 10.45$\pm$1.24
& $-2.14$
& 2.89 \\

MoME-style
& Expert routing
& 6.11$\pm$1.49
& 8.22$\pm$1.05
& 10.47$\pm$1.03
& $-2.17$
& 2.91 \\

AUAF-style
& Uncertainty gating
& 5.96$\pm$1.18
& 8.19$\pm$0.89
& 10.11$\pm$0.88
& $-2.02$
& 2.76 \\

\rowcolor{gray!15}
\addlinespace[2pt]
\textbf{GaitVista}
& \textbf{Explicit reliability cues}
& \textbf{4.31$\pm$0.93}
& \textbf{5.91$\pm$0.44}
& \underline{9.11$\pm$2.28}
& \textbf{$-0.36$}
& \textbf{1.11} \\
\midrule
Joint-frame oracle
& Ground-truth hard selector
& 3.20$\pm$0.47
& 4.96$\pm$0.41
& 6.19$\pm$0.36
& $+0.74$
& 0.00 \\
\bottomrule
\end{tabular}
}
\vspace{-5pt}
\caption{Subject-disjoint TotalCapture results under 5-pixel keypoint
perturbations. Values are mean~$\pm$~population s.d.\ across five held-out
participants after averaging five seeds and seven conditions. Worst is the
maximum condition error per participant; lower is better except
$G_{\mathrm{best}}$. Additional baselines are reported in the supplement.\vspace{-15pt}}
\label{tab:main_loso}
\end{table*}

\section{Experimental Setup}
\label{sec:experiments}

\paragraph{Datasets and protocols.}
We evaluate on TotalCapture \cite{trumble2017total} and MoVi
\cite{ghorbani2021movi} under five-fold leave-one-subject-out protocols.
TotalCapture provides controlled stress tests under reduced camera coverage,
lower-body occlusion, IMU noise, IMU dropout, and combined degradation, using
projected joints with $5$-pixel Gaussian perturbations; MoVi uses MediaPipe
keypoints from synchronized videos and a sparse optical-marker reference.
Models are trained from scratch within each dataset, held-out participants are
excluded from training and tuning, and both protocols retain
participant-specific supervised IMU mounting calibration.

\paragraph{Baselines and metrics.}
All methods receive identical pose streams and corruptions. We compare against
unimodal, fixed-fusion, filtering, feedforward, temporal, direct-regression,
expert-routing, uncertainty-gated, and entropy-gated alternatives.
Kalman-Condition and the ground-truth joint-frame selector are advantaged
diagnostics: the former receives the sensing condition, while the latter
performs optimal hard selection between streams for each joint and frame.
TotalCapture reports full-body and lower-body MPJPE, worst-condition error,
per-condition degradation, $G_{\mathrm{best}}$, and gap to the joint-frame
selector. MoVi reports error over the directly marker-supported head, wrists,
and ankles, with broader joint sets as secondary measures. Results are averaged
over five seeds and reported as mean~$\pm$~population standard deviation across
five held-out participants.

\paragraph{Downstream evaluation.}
TotalCapture predictions also drive one frozen S1 articulated Gaussian avatar
and are evaluated by bilateral knee-flexion waveform and range-of-motion error
across all five folds. Avatar geometry, appearance, and deformation are fixed
across pose methods; neither analysis contributes a training signal.

\section{Results}
\label{sec:results}

We evaluate whether \textsc{GaitVista} preserves measurement fidelity under
realistic sensing degradation, following the intended workflow: robustness,
gait-relevant kinematics, transfer to image-derived observations, and
inspection of assigned modality contributions.

\paragraph{Reliable Reconstruction under Sensing Failure.}

Table~\ref{tab:main_loso} reports subject-disjoint TotalCapture results across
seven sensing conditions. GaitVista achieves the lowest worst-condition error
among fusion methods, $9.11\pm2.28$~cm, against $10.11$ to $15.36$~cm for the
deployable alternatives shown. IMU-only is more robust in the single worst
condition at $8.26\pm1.41$~cm but has $44.1\%$ higher average error. GaitVista
therefore retains visual accuracy.

Its regret relative to the stronger complete modality is
$G_{\mathrm{best}}=-0.36$~cm, against $-2.02$ to $-2.98$~cm for the deployable
fusion baselines, and its $1.11$~cm gap to the joint-frame oracle is smaller
than their $2.76$ to $3.72$~cm gaps and the $1.78$~cm gap of Kalman-Condition,
which receives the sensing condition explicitly. GaitVista thus approaches
locally informed selection without condition labels. It also achieves the lowest average full-body and lower-body MPJPE,
$4.31\pm0.93$ and $5.91\pm0.44$~cm, reducing error by $27.7\%$ and $27.8\%$
against the strongest deployable condition-blind baseline and by $30.6\%$
against the stronger unimodal stream. It is the best deployable method in every
fold.

\begin{figure*}[!tb]
    \centering
    \includegraphics[width=0.9\textwidth]{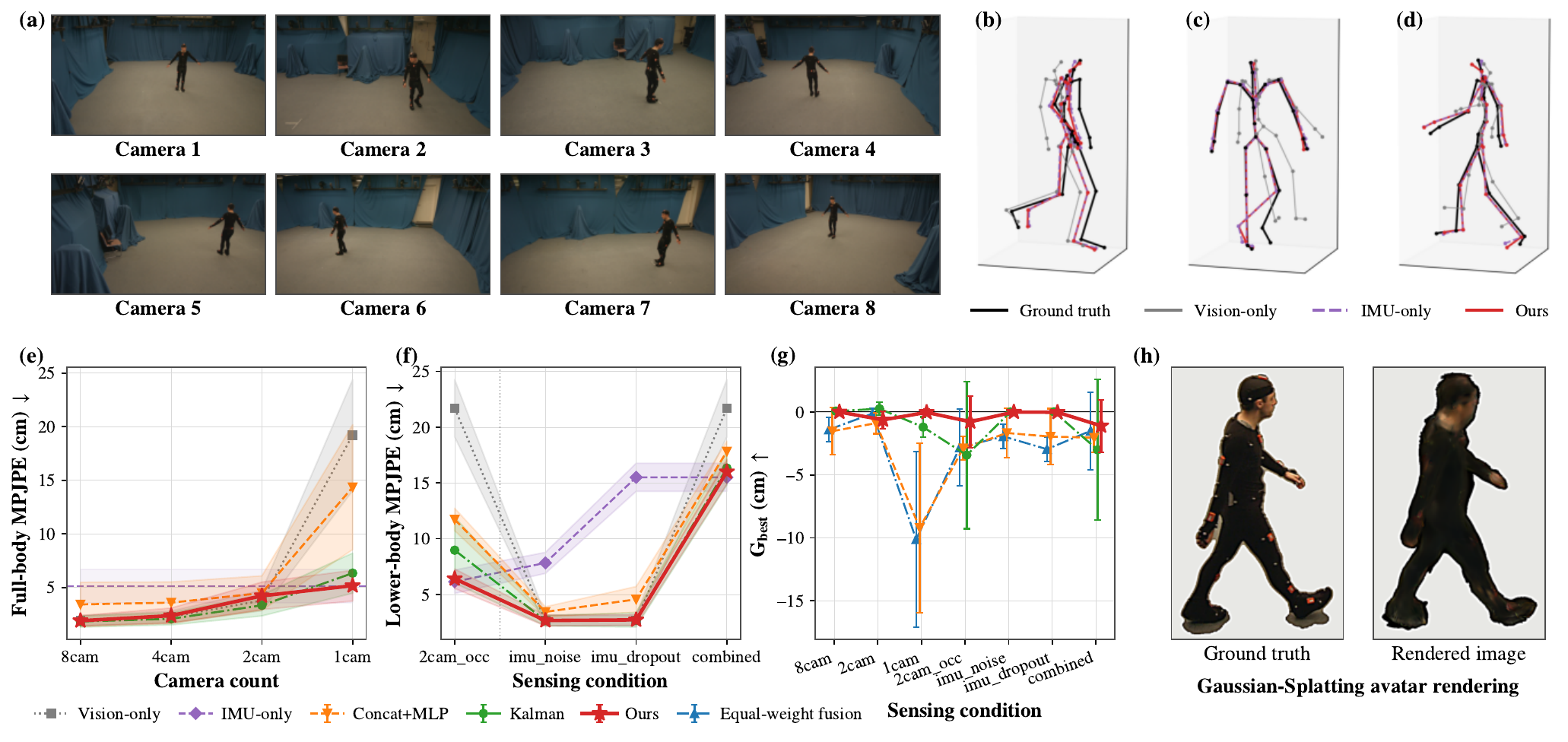}
    \vspace{-13pt}
    \caption{TotalCapture under perturbed 2D observations.
    (a) Camera setup; (b) to (d) one-camera pose comparisons;
    (e) full-body MPJPE across camera counts; (f) lower-body MPJPE under
    sensing failures; (g) gain over the stronger modality; and
    (h) fixed-avatar rendering. Curves show mean~$\pm$~population s.d.\
    across five held-out participants after seed averaging. \vspace{-15pt}}
    \label{fig:totalcapture_results}
\end{figure*}

Figure~\ref{fig:totalcapture_results} resolves performance by condition.
With eight cameras, GaitVista matches Vision-only at $1.90$~cm.  With one
camera, it reaches $5.17$~cm, close to IMU-only at $5.14$~cm and far below
Vision-only at $19.21$~cm. Under IMU noise and dropout, it remains within
$0.02$~cm of the intact visual stream. Under lower-body occlusion and combined
corruption, it improves on Vision-only but does not always surpass IMU-only;
simultaneous local failure in both modalities remains the hardest case. On an unseen four-camera configuration, the frozen model reduces error from
$3.59$ to $3.11$~cm relative to Concat+MLP, although Vision-only and
Kalman-Condition remain stronger, so transfer beyond the trained camera counts
is only partial.

\paragraph{Preservation of Gait-Relevant Kinematics.}

Rehabilitation assessment uses derived joint kinematics rather than pose error,
and those kinematics inherit the reliability of the reconstructed motion. We
evaluate bilateral knee-flexion waveform and range-of-motion agreement across
the five held-out TotalCapture participants.

GaitVista reduces condition-averaged waveform RMSE from $8.57^\circ$ to
$6.95^\circ$ relative to the strongest learned fusion baseline, an $18.9\%$
improvement, with the lowest average error for every held-out participant, and
reduces range-of-motion error from $15.67^\circ$ to $14.97^\circ$. Under
combined lower-body occlusion and IMU dropout it reaches $25.22^\circ$ RMSE,
comparable to IMU-only at $25.17^\circ$ and below the remaining fusion methods.
Pose-level robustness therefore carries into a gait-relevant signal, although
simultaneous failure in both modalities remains unresolved.

\begin{table}[!tb]
\centering
{
\small
\setlength{\tabcolsep}{3.2pt}
\begin{tabular}{@{}lccc@{}}
\toprule
Method
& Avg.\ RMSE
& Combined RMSE
& ROM error \\
& ($^\circ$) $\downarrow$
& ($^\circ$) $\downarrow$
& ($^\circ$) $\downarrow$ \\
\midrule
Vision
& 12.04$\pm$1.09
& 27.46$\pm$2.18
& 27.84$\pm$4.84 \\
IMU
& 11.08$\pm$1.03
& \textbf{25.17$\pm$1.46}
& 27.16$\pm$4.44 \\
Equal
& 10.58$\pm$0.94
& 25.74$\pm$2.45
& 25.54$\pm$4.54 \\
Concat+MLP
& 8.57$\pm$0.81
& 26.36$\pm$1.96
& 15.67$\pm$2.63 \\

\rowcolor{gray!15}
\textbf{GaitVista}
& \textbf{6.95$\pm$0.68}
& \underline{25.22$\pm$1.53}
& \textbf{14.97$\pm$2.66} \\
\bottomrule
\end{tabular}
}
\vspace{-7pt}
\caption{Bilateral knee-flexion agreement on TotalCapture under five-subject
LOSO. Combined denotes lower-body occlusion with IMU dropout. Values are
mean~$\pm$~population s.d.\ after seed averaging. \vspace{-18pt}}
\label{tab:downstream}
\end{table}

Published minimal clinically important differences for sagittal knee
range of motion after stroke are smaller than the remaining error reported
here. The comparison serves only as a scale reference: the evaluation uses
neurologically healthy participants and a segment-based 3D knee angle rather
than a clinically calibrated sagittal measure, and identifies a remaining
accuracy gap rather than clinical readiness.

\paragraph{External Validation with Image-Derived Keypoints.}

To complement TotalCapture's perturbed projections and protocol-supplied
occlusion indicator, we evaluate on MoVi with MediaPipe keypoints from two
synchronized videos, training from scratch within MoVi and excluding held-out
participants from development.

Vision is far stronger than IMU on MoVi: $12.00$ versus $49.70$~cm Marker-5
MPJPE. Equal fusion raises error to $31.64$~cm, and both learned baselines
remain worse than Vision-only. GaitVista instead achieves
$11.64\pm0.99$~cm, improving over the strongest learned fusion baseline by
$6.4\%$ and Vision-only by $3.0\%$. It outperforms both learned baselines on
all five held-out participants and Vision-only on four, trailing by only
$0.12$~cm on the fifth.

GaitVista is also the only deployable fusion method with positive mean
$G_{\mathrm{best}}$ $(+0.36$~cm$)$. Its mean visual contribution is $0.953$,
indicating conservative recalibration around the stronger visual stream rather
than binary modality switching. The learned reliability rule therefore avoids
overusing a weak modality under image-derived observations.

\begin{table}[!tb]
\centering
{
\small
\setlength{\tabcolsep}{1.5pt}
\begin{tabular}{@{}lcccc@{}}
\toprule
Method
& \multicolumn{3}{c}{MPJPE (cm) $\downarrow$}
& $G_{\mathrm{best}}^{\mathrm{M5}}$ \\
\cmidrule(lr){2-4}
& Marker-5
& Ankle-2
& Full-24
& (cm) $\uparrow$ \\
\midrule
Vision
& 12.00$\pm$0.81
& 12.61$\pm$1.54
& 11.39$\pm$1.09
& -- \\

IMU
& 49.70$\pm$3.66
& 26.78$\pm$1.36
& 32.99$\pm$3.21
& -- \\

Equal
& 31.64$\pm$1.58
& 23.50$\pm$2.48
& 22.09$\pm$1.62
& $-19.64$ \\

Concat+MLP
& 12.43$\pm$1.05
& 12.61$\pm$1.56
& 11.59$\pm$1.27
& $-0.43$ \\

MoME
& 12.56$\pm$0.92
& 13.55$\pm$1.55
& 11.60$\pm$1.23
& $-0.56$ \\

\rowcolor{gray!15}
\textbf{GaitVista}
& \textbf{11.64$\pm$0.99}
& \textbf{12.30$\pm$1.97}
& \textbf{11.20$\pm$1.13}
& \textbf{+0.36} \\

\midrule
Joint-frame oracle
& 11.03$\pm$0.66
& 11.44$\pm$1.16
& 9.99$\pm$1.12
& +0.97 \\
\bottomrule
\end{tabular}
}
\vspace{-7pt}
\caption{Subject-disjoint MoVi results using MediaPipe keypoints from two synchronized views. Marker-5 and Ankle-2 are directly marker-supported; Full-24 partly relies on sparse-marker fitting. Values are mean~$\pm$~population s.d.\ over five held-out participants after seed averaging.\vspace{-15pt}}
\label{tab:movi}
\end{table}

\begin{figure*}[!tb]
    \centering
    \includegraphics[width=\textwidth]{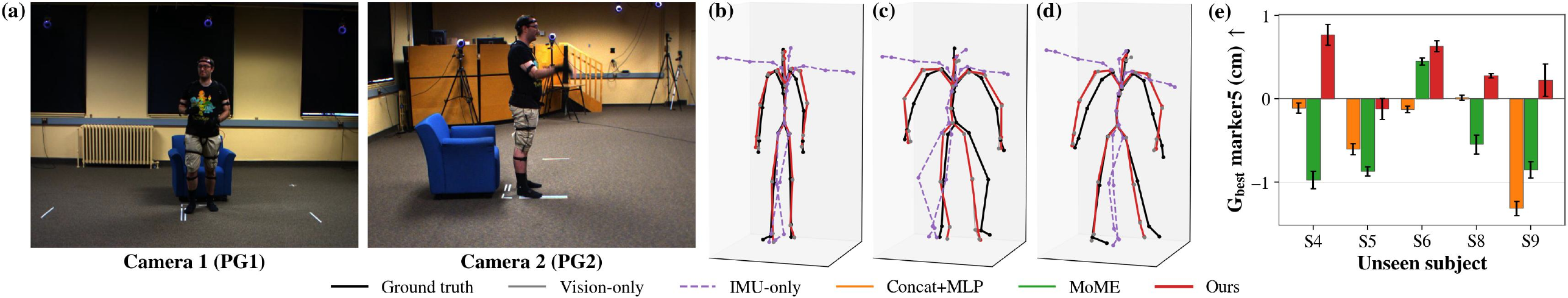}
    \vspace{-19pt}
    \caption{MoVi evaluation with image-derived keypoints.
    (a) Synchronized views; (b) to (d) held-out pose comparisons; and
    (e) participant-wise $G_{\mathrm{best}}$ over marker-supported joints.
    Positive values indicate improvement over both unimodal streams; error
    bars show variation across five seeds. \vspace{-13pt}}
    \label{fig:movi_results}
\end{figure*}

Figure~\ref{fig:movi_results} shows the input views, held-out reconstructions,
and participant-wise fusion gain. GaitVista attains positive
$G_{\mathrm{best}}$ for four of five participants, while the learned baselines
are predominantly negative.

\paragraph{Reliability Diagnostics and Ablations.}

Table~\ref{tab:ablations} tests whether fixed anatomical preference suffices.
Retaining only the learned joint embedding and removing all frame-varying cues
increases average MPJPE by $2.74$~cm and lower-body MPJPE by $4.42$~cm, so
location alone cannot capture reliability changes within a trial. Cross-modal
disagreement and root continuity provide complementary evidence: removing
disagreement increases average error by $0.38$~cm, and removing root continuity
increases it by $0.72$~cm and raises lower-body error from $5.91$ to
$6.96$~cm. Every ablation also worsens performance under combined visual and
inertial degradation. Reliability therefore requires anatomical priors together
with frame-varying sensing cues.

\begin{table}[!tb]
\centering
{
\small
\setlength{\tabcolsep}{2.0pt}
\begin{tabular}{@{}lcccc@{}}
\toprule
Variant
& \multicolumn{3}{c}{MPJPE (cm) $\downarrow$}
& $\Delta$ Avg. \\
\cmidrule(lr){2-4}
& Avg.
& Lower
& Combined
& (cm) $\downarrow$ \\
\midrule
Static joint
& 7.05$\pm$0.99
& 10.33$\pm$2.01
& 9.53$\pm$0.84
& $+2.74$ \\

No disagreement
& 4.69$\pm$0.80
& 6.47$\pm$1.15
& 9.18$\pm$2.23
& $+0.38$ \\

No root continuity
& 5.03$\pm$1.06
& 6.96$\pm$2.00
& 9.43$\pm$1.88
& $+0.72$ \\

No visual quality 
& 6.60$\pm$1.06
& 9.24$\pm$1.90
& 9.39$\pm$0.86
& $+2.29$ \\

\rowcolor{gray!15}
\textbf{GaitVista}
& \textbf{4.31$\pm$0.93}
& \textbf{5.91$\pm$0.44}
& \textbf{9.11$\pm$2.28}
& \textbf{0.00} \\
\bottomrule
\end{tabular}
}
\vspace{-5pt}
\caption{TotalCapture ablations with 5-pixel keypoint perturbations. Values are mean~$\pm$~population s.d.\ across five held-out participants after seed averaging. Avg.\ and Lower average seven conditions; Combined denotes lower-body occlusion with IMU dropout. $\Delta$ Avg.\ is relative to GaitVista, and Static joint uses only joint-specific preference. \vspace{-18pt}}
\label{tab:ablations}
\end{table}

Assigned contributions indicate how the model combines modalities, not
calibrated correctness. Some joints remain strongly vision-weighted under
monocular depth ambiguity, so the weights support inspection of modality use
without replacing review of the reconstructed motion.

\paragraph{Interpretation and Inspection.}

\begin{figure}[!tb]
    \centering
    \makebox[\columnwidth][c]{%
        \begin{minipage}[t]{0.43\columnwidth}
            \centering
            \includegraphics[width=\linewidth]{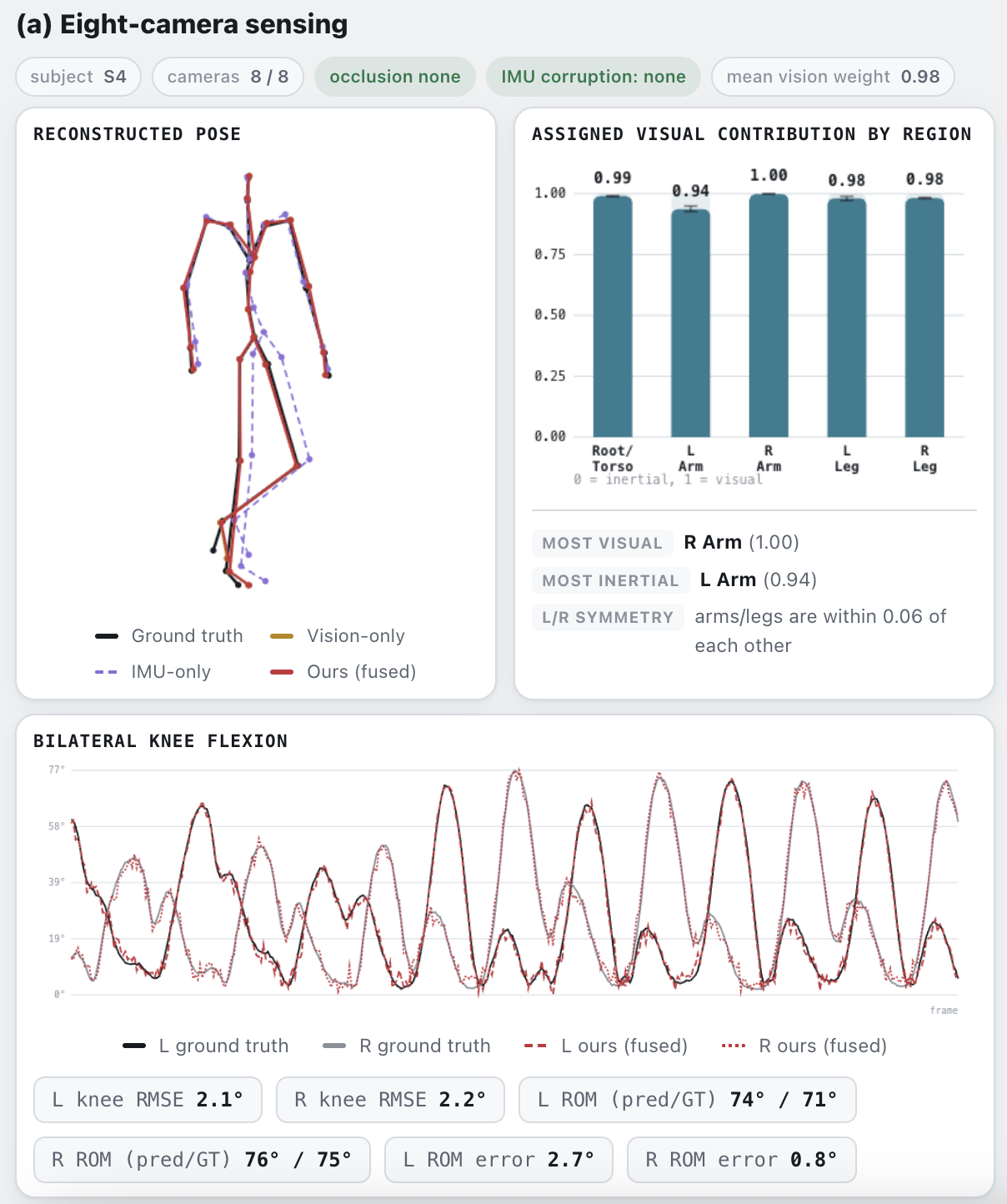}
        \end{minipage}%
        \hspace{0.02\columnwidth}%
        \begin{minipage}[t]{0.43\columnwidth}
            \centering
            \includegraphics[width=\linewidth]{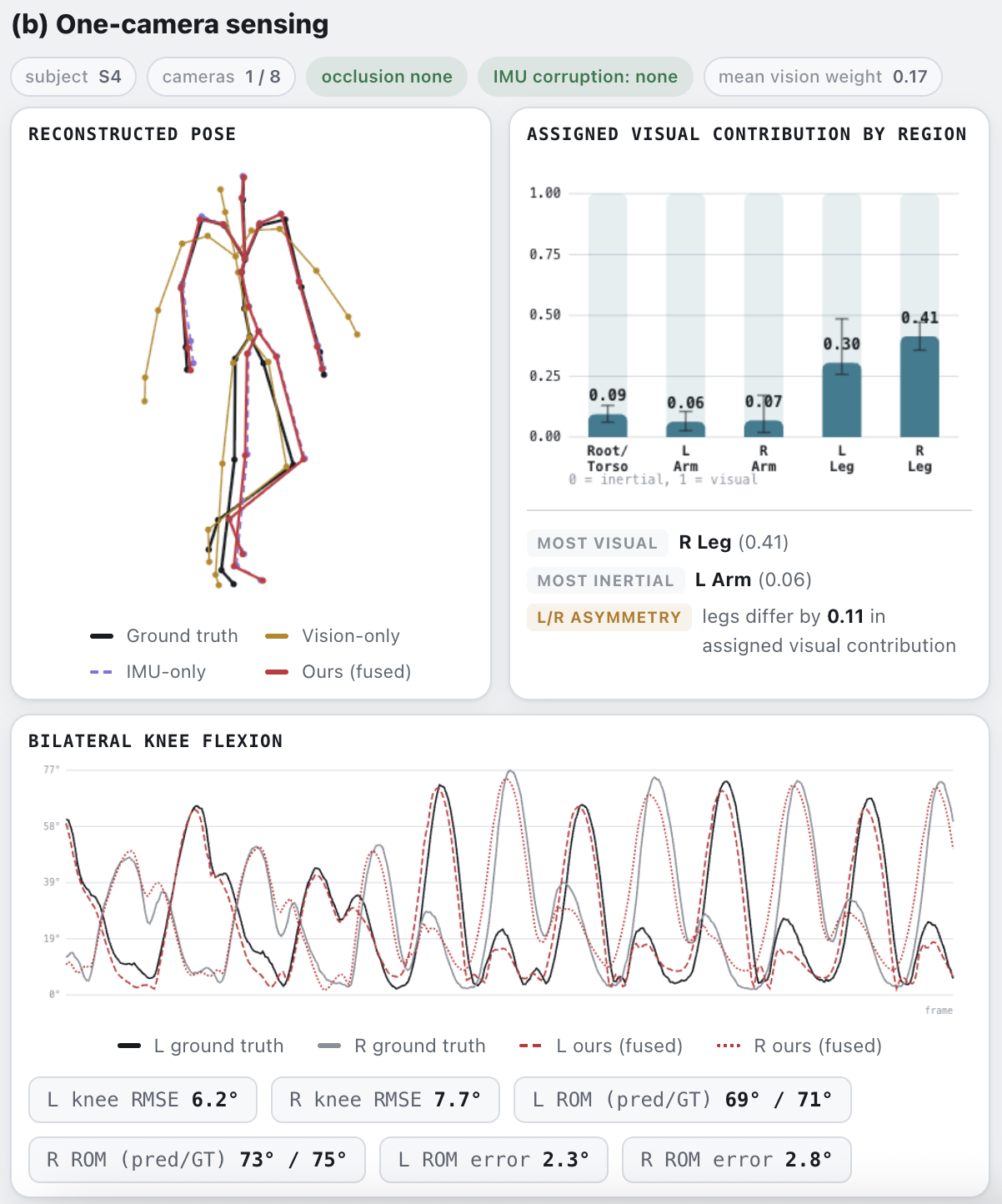}
        \end{minipage}%
        }
    \vspace{-15pt}
    \caption{Inspection prototype for one held-out sequence with
(a) eight-camera and (b) one-camera sensing. Views show reconstructed pose,
regional visual contributions, and bilateral knee-flexion trajectories.
Contributions reflect modality use, not calibrated correctness. \vspace{-18pt}}
    \label{fig:inspection}
\end{figure}

Figure~\ref{fig:inspection} shows a lightweight prototype combining the
reconstructed motion, assigned modality contributions, and bilateral
knee-flexion trajectories for a held-out sequence; it loads frozen
subject-disjoint outputs and performs no additional inference. Driving the same
fixed S1 Gaussian avatar with every pose method, GaitVista also achieves the
best PSNR, SSIM, and LPIPS by small margins, with full rendering results in the
supplement. Neither the avatar views nor the interface has been evaluated for
clinical usability.

\section{Clinical Relevance and Limitations}
\label{sec:clinical_relevance}

Post-stroke gait assessment commonly uses speed, cadence, bilateral symmetry,
and lower-limb excursion. Asymmetry relates to balance, fall risk, and stroke
severity \cite{patterson2008asymmetry,xu2025asymmetry}, and knee range of
motion has reported clinically important differences \cite{guzik2020knee}.
These measures are derived from reconstructed motion and inherit its errors;
unequal sensing quality across limbs can create apparent asymmetry. The
robustness metrics characterize that failure mode under the tested conditions,
and the assigned contributions expose shifts in modality use.

Both benchmarks contain healthy participants in controlled settings, so
performance on pathological gait, assistive-device use, and clinical outcomes
is not established. The protocols retain supervised participant-specific IMU
calibration, TotalCapture supplies the occlusion indicator, and shared root
translation excludes speed and step-length evaluation. MoVi has five held-out
participants and sparse marker support beyond the head, wrists, and ankles. The
interpretation study uses one frozen avatar without clinician or patient
evaluation, simultaneous visual and inertial failure remains unresolved, and
the raw inertial analysis measures magnetic disturbance rather than pose error.
Future work should address pathological gait, image-inferred reliability,
reduced calibration, global trajectory, and clinical outcomes.

\section{Conclusion}
\label{sec:conclusion}

Changes in sensing quality can mimic movement changes, limiting accessible gait
assessment. \textsc{GaitVista} uses reliability-aware vision--IMU fusion with
inspectable joint-wise contributions. On TotalCapture, it achieves the lowest
average full- and lower-body errors, best worst-condition fusion result,
smallest selector gap, and near-zero stronger-modality regret. On MoVi, it
preserves the stronger visual estimate, with knee-flexion gains extending to
gait-relevant kinematics. Pathological gait and clinical validation remain
open.

\bibliography{references}

\end{document}